\documentclass[letterpaper, 10 pt, conference, final]{ieeeconf}
\usepackage{mathtools}
\usepackage{amsmath}
\usepackage{amssymb}
\usepackage{xcolor}
\usepackage[noadjust,sort]{cite}
\usepackage[font=footnotesize]{caption}
\usepackage{subcaption}
\usepackage{balance}
\usepackage{soul}

\IEEEoverridecommandlockouts
\title{\LARGE \bf Learning Periodic Tasks from Human Demonstrations}

\author{Jingyun Yang$^{1\dagger}$, Junwu Zhang$^{2}$, Connor Settle$^{3}$, Akshara Rai$^{4}$, Rika Antonova$^{3\ddagger}$, Jeannette Bohg$^{3}$ % <-this % stops a space
\thanks{1. Machine Learning Department, Carnegie Mellon University; 2. Department of Mechanical Engineering, Stanford University; 3. Department of Computer Science, Stanford University; 4. Meta AI.}%
\thanks{$\dagger$ Work done while the author was an intern at IPRL lab at Stanford.}%
\thanks{$\ddagger$ Supported by the National Science Foundation grant No.2030859 to the Computing Research Association for the CIFellows Project.}%
\thanks{\scriptsize{Contacts: \texttt{jingyuny@andrew.cmu.edu}, \texttt{rika.antonova@stanford.edu}}}%
}

\newcommand{\vc}[1]{\boldsymbol{#1}}

\newcommand{\Dc}{\mathcal{D}}

\newcommand{\lv}{\mathbf{l}}

\newcommand{\vv}{\mathbf{v}}
\newcommand{\wv}{\mathbf{w}}

\newcommand{\name}{ViPTL}

\newcommand{\edit}[1]{#1}

\begin{document}

\maketitle
\thispagestyle{empty}
\pagestyle{empty}

\begin{abstract}
We develop a method for learning periodic tasks from visual demonstrations. The core idea is to leverage periodicity in the policy structure to model periodic aspects of the tasks. We use active learning to optimize parameters of rhythmic dynamic movement primitives (rDMPs) and propose an objective to maximize the similarity between the motion of objects manipulated by the robot and the desired motion in human video demonstrations. We consider tasks with deformable objects and granular matter whose \edit{states are} challenging to represent and track: wiping surfaces with a cloth, winding cables, \edit{and} stirring granular matter with a spoon. Our method does not require tracking markers or manual annotations. The initial training data consists of 10-minute videos of random unpaired interactions with objects by the robot and human. We use these for unsupervised learning of a keypoint model to get task-agnostic visual correspondences. Then, we use Bayesian optimization to optimize rDMPs from a single human video demonstration within few robot trials. We present simulation and hardware experiments to validate our approach.
\end{abstract}
\section{INTRODUCTION}

\begin{figure}
    \centering
    \includegraphics[width=0.475\textwidth]{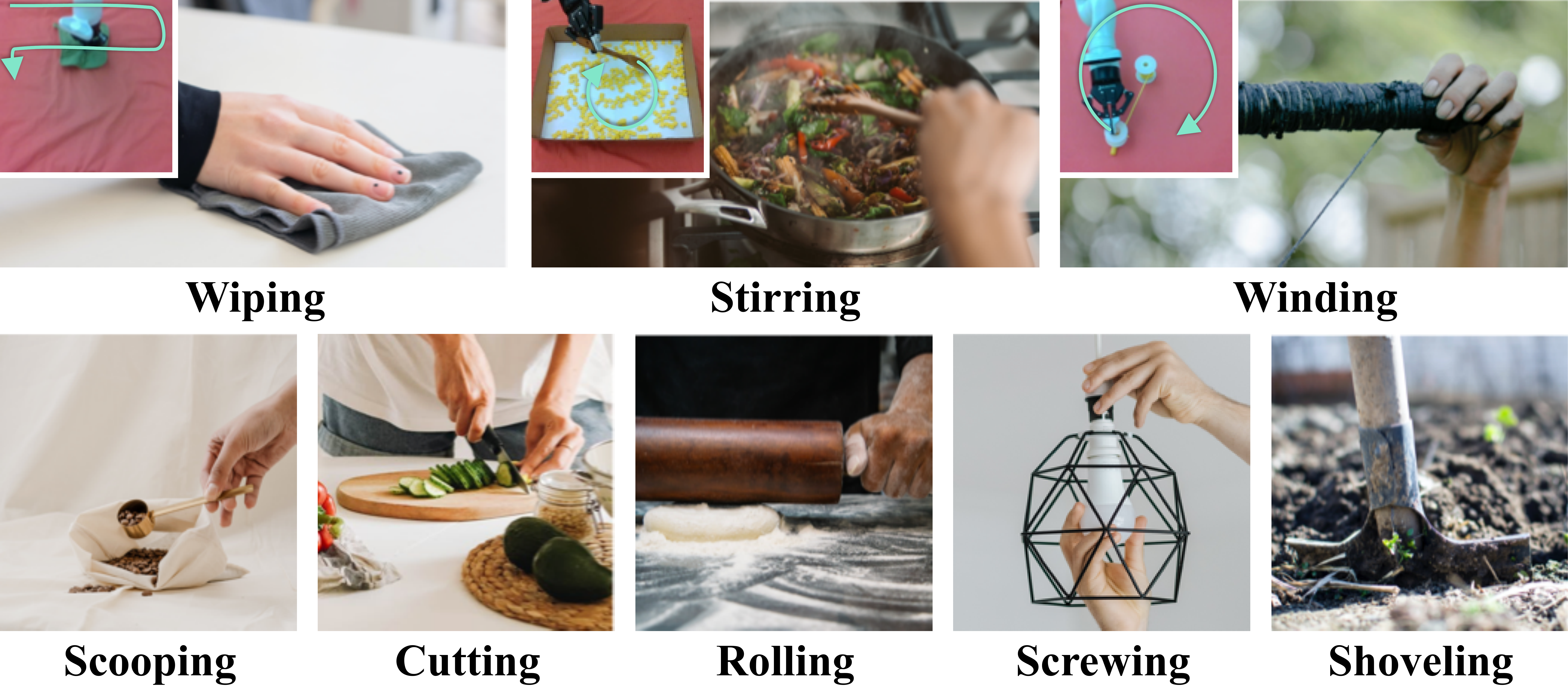}
    \caption{Examples of everyday periodic tasks. Our robot learns to imitate wiping, stirring and winding from human video demonstrations (top row).}
    \label{fig:teaser}
\end{figure}

Periodic tasks such as wiping a table with a cloth, stirring food, winding cables, or tying ropes are ubiquitous in our daily lives (see Figure~\ref{fig:teaser}).
In this work, we address how robots can appropriately represent and learn periodic policies by watching humans. While prior works considered learning manipulation skills from human demonstrations~\cite{liu2018imitation, smith2019avid, sharma2019third, xiong2021learning, shao2020concept, chen2021learning}, less attention has been given to periodic tasks. These tasks repeat similar motion with only small differences between repetitions. If a robot was able to decompose the demonstration of a periodic task into periods, it could efficiently learn the underlying motion and repeat it as many times as necessary. Prior hierarchical learning approaches investigated learning compositional tasks either from demonstrations \cite{xu2018neural, Silver2019FewShotBI} or through reinforcement learning \cite{Nair2020HierarchicalFS, hundt2020good}. However, these approaches do not leverage the strong relationship \edit{across} repetitions and can be inefficient for learning periodic tasks.

In this work, we propose {\em Visual Periodic Task Learner\/} (\name): a method with an explicit periodic policy representation that enables efficient learning of periodic robot manipulation skills from a single, visual human demonstration. \edit{For} policy representation\edit{,} we \edit{adopt} rhythmic {\em Dynamic Movement Primitives\/} (rDMPs) \cite{Ijspeert2002LearningRM} which model cyclic motion, with shift parameters that account for motion that translates over multiple periods. The benefits of using rhythmic DMPs are two-fold: (1) they succinctly represent periodic manipulation policies, enabling efficient learning -- the problem of learning a long-horizon periodic task is reduced to learning a single-period motion of the task; (2) the learned rhythmic DMP can be repeated any number of times and the speed and amplitude of the motion can be adjusted at run time by a simple change of a parameter.

Typically, DMPs are trained on robot trajectories demonstrated through kinesthetic teaching or teleoperation. However, this can be non-intuitive to non-expert users. Instead, we use human video demonstrations that do not contain trajectories but are easier to record. To train rDMPs from such videos, we learn a keypoint detector that identifies consistent keypoints across human demonstrations and robot executions. Based on this, we can now evaluate the similarity between a robot execution and the human demonstration -- a quantity we aim to maximize. Keypoint models do not assume rigid objects and are therefore well suited for challenging manipulation tasks considered in this paper that involve deformable objects and granular material. We learn the keypoint model from task-agnostic play data and leverage periodicity estimation from~\cite{dwibedi2020counting} to break down the demonstrated task into single-period components. We use Bayesian optimization (BO) to optimize similarity of the robot motion to the human demonstration. 
To focus BO on promising regions, we create `imagined' trajectories from segments of robot play data that serve as initial  candidates for BO.

\begin{figure*}[ht]
\vspace{6pt}
    \centering
    \includegraphics[width=0.98\textwidth]{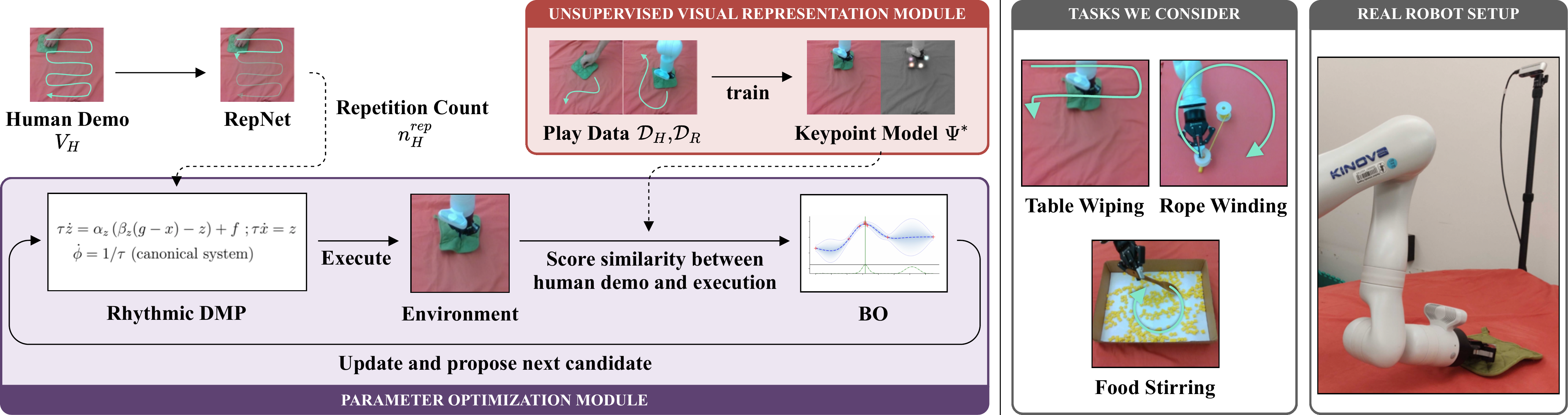}
    \caption{Overview of the proposed approach. Our method is composed of two modules: the unsupervised visual representation module and the parameter optimization module. The unsupervised module learns a model for keypoint correspondences between the motions of the objects in the human demonstration and the robot trials. The parameter optimization module uses active learning to adjust the parameters of the rhythmic DMP controllers.}
    \label{fig:method_overview}
\end{figure*}

We quantitatively evaluate the proposed method in both simulation and on a real robot with three manipulation tasks: table wiping, rope winding, and food stirring (see Figure~\ref{fig:teaser}). We show that our approach can successfully learn challenging periodic manipulation tasks that involve deformable and granular objects from a single human demonstration within 50 robot trials. Our comparisons to existing approaches and ablations show how our perception and optimization modules contribute to the overall success of the method.

\edit{In summary, our contributions are as follows: (1) a framework that represents and learns periodic robot manipulation policies from human demonstrations; (2) a motion optimization algorithm that leverages BO for sample-efficient learning and an unsupervised keypoint model for score computation in BO; (3) empirical results on a range of periodic manipulation tasks that involve deformable and granular objects.}
\section{RELATED WORK}

\subsection{Modeling Periodic Motion}

Various methods have been proposed to model periodic motion. Early literature in robotics and neuroscience used limit cycles and central pattern generators to model periodic motions for locomotion~\cite{jalics1997pattern, kajita2002realtime, yang2004infant,2008ICRALudo}. Recently, pattern generators have also been used with robotic manipulators \cite{thor2019fast, oikonomou2020periodic}. 
However, limit cycle formulations are not easily amenable to learning arbitrary periodic trajectories.
In such cases, dynamic movement primitives (DMPs)~\cite{Ijspeert2002LearningRM} can provide the needed flexibility, and ease of use with learning-based approaches. DMPs have been used for periodic manipulation tasks, writing and wiping being the most common~\cite{6651499, 10.1162/NECO_a_00393}. In comparison, our proposed work also learns winding and stirring tasks, but from visual demonstration that are not annotated with human hand poses. {\em Fourier movement primitives\/} (FMPs)~\cite{kulak2020fourier} are an extensions of DMPs using Fourier series as basis functions.
While our system is agnostic to the specific choice of periodic parameterization of the control policies, here we use rhythmic DMPs. 

\subsection{Periodicity Estimation}

There has been a significant interest in estimating periodicity in the computer vision community. Prior works use Fourier analysis~\cite{azy2008segmentation, cutler2000robust, pogalin2008visual}, singular value decomposition~\cite{chetverikov2006motion}, or peak detection~\cite{thangali2005periodic} to detect repetition by converting the motion in videos to one-dimensional signals.
Recent works propose detecting non-stationary repetitive motion using wavelet transforms~\cite{runia2018real}, 3D convolution networks~\cite{zhang2020context}, and self-similarity between video frames~\cite{dwibedi2020counting}.
In this work, we use RepNet~\cite{dwibedi2020counting} for periodicity estimation. We find that once trained on the Countix dataset in \cite{dwibedi2020counting}, RepNet can successfully decompose human demonstrations of various manipulation tasks into single-period segments without further finetuning.

\subsection{Learning from Human Demonstrations}
Several works in learning from human video demonstrations propose using image-to-image translation to transform human demos to robot executions~\cite{liu2018imitation, smith2019avid, sharma2019third, xiong2021learning}. However, these require a large amount of training data. Recent works~\cite{shao2020concept, chen2021learning} leverage action recognition models, such as the action classifiers trained on the 20BN Something-Something dataset~\cite{goyal2017something}, to identify whether the robot is performing the desired task. However, while these classifiers are useful for identifying the class of motions for short interactions, we show that they do not retain enough information to analyze tasks with longer duration and multiple repetitions.

Our approach uses a small amount of task-agnostic, unpaired and unlabeled `play' data~\cite{lynch2020learning} to train a keypoint model that makes it possible to quantitatively compare human demos and robot executions. `Play' data is useful, because it can be collected without supervision and using a task-agnostic, randomized policy. However, unlike~\cite{lynch2020learning} that explores using hours of such data, we focus on a much more data-efficient alternative. We collect 10 minutes of human `play' data and 10 minutes of robot `play' data (unpaired), and then use a single human demonstration video to infer the appropriate parameters for the robot control policy. 

\subsection{Sample-efficient Robot Learning}

We aim to imitate a visual human demo on a robot with high sample-efficiency. Prior works have explored model-based methods~\cite{hafner2020mastering, ebert2018visual, sekar2020planning} and self-supervised exploration algorithms~\cite{pathak2017curiosity, schmidhuber1991possibility, klyubin2005empowerment, eysenbach2018diversity, bellemare2016unifying, yarats2021reinforcement} to improve sample efficiency, but these methods often require much more data on the robot and do not directly generalize to visual imitation learning. Some works utilize large-scale training in simulation and transfer the learned policies to the real robot~\cite{tobin2017domain, james2019sim}. Since our tasks include hard-to-simulate deformable and granular objects, sim2real is not applicable in our setup. In contrast, we use Bayesian Optimization (BO) to optimize parameters of a rhythmic DMP. BO is capable of learning the demonstrated manipulation skills within 50 trials on the real robot.
\section{PRELIMINARIES} \label{sec:prelim}

We consider the problem of learning periodic manipulation skills from a single human demonstration. We assume that the human demonstrates a periodic task that is performed for at least 2 periods. Given a human demonstration $V_H = (I_H^{1}, \ldots, I_H^{T_H})$ as a sequence $V_H$ of $T_H$ RGB image frames $I_H^{t}$, the robot is allowed to execute a total of 50 trials in the environment to learn the demonstrated skill\edit{, after which the best execution is selected for evaluation}. In each trial, the robotic agent executes a trajectory \edit{$\tau_R = (\vc{x}_R^{1}, \ldots, \vc{x}_R^{T})$ where $\vc{x}^t$} denotes the robot end-effector position at timestep $t$. We denote the corresponding execution video of this executed trajectory as $V_R = (I_R^{1}, \ldots, I_R^{T})$, where $I_R^t$ denotes the camera image at timestep $t$. 
We measure the similarity between the object motion in the human demonstration and in the video of the robot execution using consistent keypoints across the two videos (see Section~\ref{method:keypoint}).
In this work, we combine BO and DMPs to maximize this similarity score, and present our method in Section~\ref{sec:method}. Below we provide the technical background for BO and DMPs.

\subsection{Bayesian Optimization}

In Bayesian optimization (BO), an optimization problem is viewed as finding parameters $\wv$ that optimize some objective function $f_{\text{BO}}(\wv): f_{\text{BO}}(\wv^*) = \max_\wv f_{\text{BO}}(\wv)$. $f_{\text{BO}}$ is commonly modeled with a Gaussian process (GP): $f_{\text{BO}}(\wv) \sim \mathcal{GP}(m(\wv), k(\wv_i, \wv_j))$. At each trial, to select the next promising candidate $\wv$, BO optimizes an acquisition function, e.g. the Upper Confidence Bound (UCB)~\cite{srinivas2009gaussian}, which explicitly balances exploration (high posterior uncertainty) \edit{vs.} exploitation (high posterior mean estimate): $UCB(\wv) = m(\wv) + \edit{\beta\cdot Std(\wv)}$.
The kernel defines a similarity function on the search space. The RBF kernel is a common choice: $k(\wv_i, \wv_j) = \sigma^2_k \exp(-\frac{1}{2}\lVert \wv_i \text{-} \wv_j \rVert_2^T \text{diag}(\lv)^{-2} \lVert \wv_i \text{-} \wv_j \rVert_2)$, where $\sigma^2_k$ and $\lv$ are signal variance and a vector of length scales, respectively. In practice, $\sigma^2_k$ is a hyperparameter optimized automatically by maximizing marginal likelihood.

\begin{figure}
\vspace{6pt}
    \centering
    \includegraphics[width=0.44\textwidth]{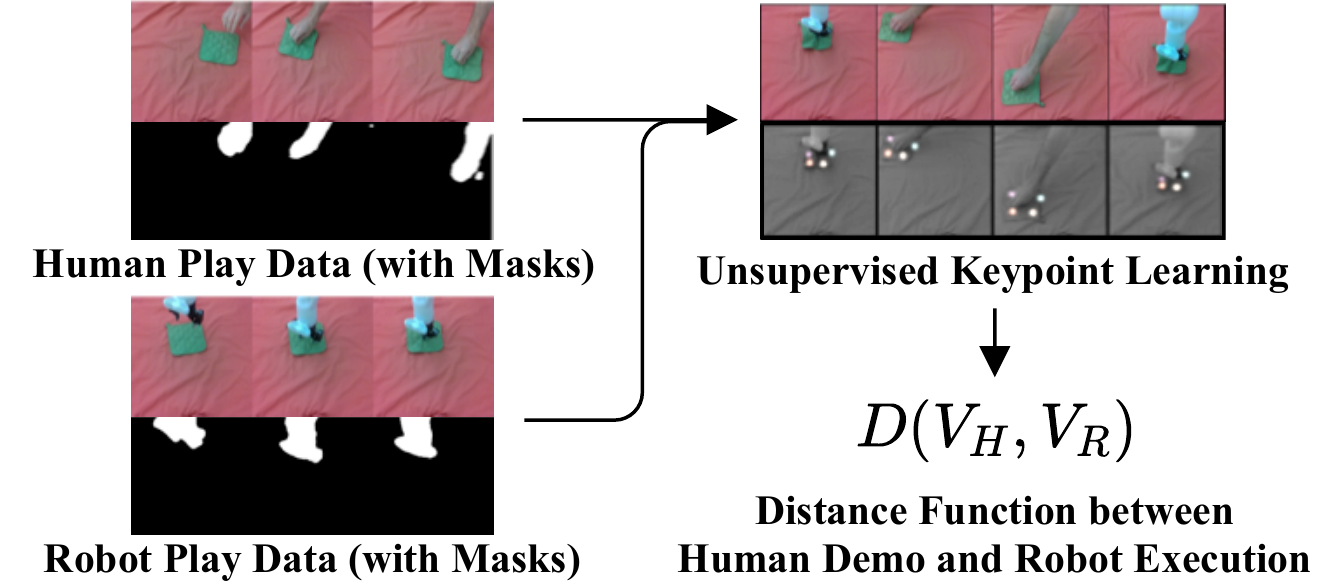}
    \caption{Unsupervised keypoint learning from play data. The play data consists of unpaired, unlabeled and task-agnostic human and robot motion recorded from the same viewpoint. It is used to train a keypoint model similar to \cite{kulkarni2019unsupervised} that finds consistent keypoints across human and robot demonstrations. This allows to compute a keypoint-based distance between human and robot videos suitable for guiding the search for the best matching robot motion. Robot trajectories are included in the robot play data, but they are not used for keypoint learning purposes.}
    \label{fig:keypoint}
\end{figure}

\subsection{Dynamic Movement Primitives (DMPs)}
DMPs are trajectory generators whose parameters can be learned from demonstrations of desired robot end-effector trajectories. They combine linear fixed-point attractors with non-linear function approximators to encode complex trajectories, while maintaining convergence guarantees. We refer readers to \cite{ijspeert2013dynamical} for a detailed overview.

The \textit{transformation system} of a DMP consists of a damped linear feedback term, and a forcing function $f$: 
\begin{equation}\label{eq:base_dynamical_system}
\begin{aligned}
\tau \dot{z} &= \alpha_{z}\left(\beta_{z}(g-x)-z\right) + f \ ; \quad \tau \dot{x} = z \ ,
\end{aligned}
\end{equation}
where $x$ is position, $g$ is the goal, $\alpha_z$ and $\beta_z$ are constants, $\tau$ is a temporal scaling factor, $z$ is the scaled velocity, and the output of the transformation system is the scaled acceleration $\dot{z}$.
The second component of DMPs is a \textit{canonical system} which replaces time, and enables scaling the trajectory to different time lengths. The canonical system is different between rhythmic and discrete DMPs; in the discrete case it represents `time left' and goes to 0 at the end of the motion, while in the rhythmic case it represents the time from the start, and goes up linearly. Specifically, in rhythmic DMPs, the first-order canonical system $\tau \dot{\phi}=1$ encodes the phase $\phi$, increasing linearly as motion progresses.

In rhythmic DMPs, the forcing function $f$ is parameterized by $\phi$ and consists of cyclic basis functions:
\begin{equation}\label{eq:rdmp_def}
\begin{aligned}
f &=\frac{\sum_i \Psi_{i} w_{i}}{\sum_i \Psi_{i}} r\ ;
\
\Psi_{i} \!=\! \exp \left(h_{i}\left(\cos \left(\phi-c_{i}\right)-1\right)\right),
\end{aligned}
\end{equation}
where $\Psi$ is a function of the canonical system, and the weights $w_i$ are commonly learned using locally weighted regression~\cite{schaal1998constructive}. The cyclic nature of basis functions ensures that the transformation system yields cyclic motion, as the canonical system unrolls. 
Typically, the goal $g$ of a rhythmic DMP is set to the mean of the demonstration trajectory and kept fixed. Discrete DMPs are shown to generalize well to changing goals. \cite{ijspeert2013dynamical} present ways to continuously change goals to new locations without causing a discontinuity in the acceleration $\dot{z}$. We adapt \cite{ijspeert2013dynamical} to smoothly move the goals of rhythmic DMPs between executions. This continuously modulates the mean point of the limit cycle of the DMP, allowing us to model motions that are mostly cyclic, but slightly shifting over time, e.g. as in wiping a surface.
\section{METHOD} \label{sec:method}

Our framework is composed of two parts: (1) a representation learning module, where a keypoint detection model is trained to extract consistent keypoints from independently collected and non-task-specific human and robot play data; (2) a parameter optimization module, where BO searches for a rhythmic DMP that \edit{would} produce a robot video that matches the human demo in terms of the detected keypoints. These modules are detailed in Sections \ref{method:keypoint} and \ref{method:bo}, respectively. Figure \ref{fig:method_overview} shows an overview.

\subsection{Unsupervised Keypoint Learning from Play Data}
\label{method:keypoint}

\begin{figure}[t]
\vspace{6pt}
    \centering
    \includegraphics[width=0.46\textwidth]{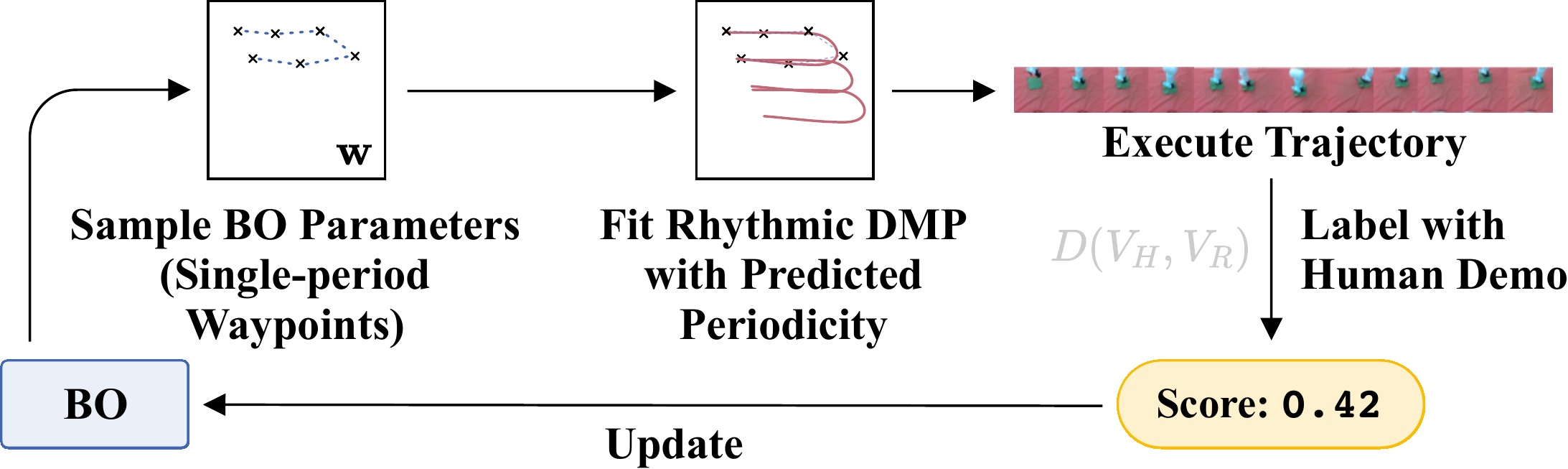}
    \caption{Our Bayesian optimization pipeline.}
    \label{fig:bo}
\end{figure}

To learn a manipulation skill from a human demo, we need a way to evaluate the similarity between the demo and robot execution. To learn such a similarity score, we assume that the agent has access to a small amount of human and robot play data. Play data is a dataset of self-guided, task-agnostic, and diverse interactions.
The human play data $\Dc_H = (I_{HP}^1, \ldots, I_{HP}^{T_{HP}})$ is a sequence of $T_{HP}$ unlabeled RGB image frames, while the robot play data is a sequence of $T_{RP}$ RGB image frames $\Dc_R = (I_{RP}^1, \ldots, I_{RP}^{T_{\edit{RP}}})$ accompanied by robot end-effector positions \edit{$\vc{x}_{RP}^{1}, \ldots, \vc{x}_{RP}^{T_{RP}} \in \mathbb{R}^3$}. Note that $\Dc_H$ and $\Dc_R$ are unpaired and independent.

To acquire a visual representation for the manipulated object that is invariant to change of agent between human and robot, we adopt a variation of the Transporter architecture \cite{kulkarni2019unsupervised} -- an unsupervised keypoint detection model that learns to generate temporally consistent keypoints $\Psi^*(I)$ on image input $I$. The learning process is illustrated in Figure \ref{fig:keypoint}. Humans and robots may move their hands very differently to generate the same object movement. To make sure that the keypoint model $\Psi^*$ allocates keypoints to the manipulated objects and not to the human and robot hand, we mask these areas in the reconstruction loss~\cite{kulkarni2019unsupervised} when training the keypoint model. This is achieved by using commonly available hand detectors~\cite{lugaresi2019mediapipe} and depth filtering, respectively, when computing the reconstruction loss that the Transporter is trained on. With this, the keypoint model is more likely to place keypoints on the objects, and be robust to different visual appearance of human and robot hands.

After the keypoint model is trained, we process the human demo $V_H$ and robot execution $V_R$ to produce sub-sampled videos $V_{H^{'}} = \{I_{H'}^i\}_{i=1}^{N_s}$ and $V_{R^{'}} = \{I_{R'}^i\}_{i=1}^{N_s}$ that both have length $N_s$. We then define the distance between the human demo and the robot execution as:
\vspace{-0.1cm}
\begin{align}
D(V_H, V_R) \!=\! \frac{1}{N_s N_k} \sum_{i=1}^{N_s} \lVert \Psi^*(I_{H^{'}}^i) - \Psi^*(I_{R^{'}}^i) \rVert_1,
\end{align}
where $\lVert \cdot \rVert_1$ denotes L$_1$ norm and $\Psi^*(I)$ denotes the locations of the $N_k$ detected keypoints of an image $I$ normalized to range $[0, 1]$. Note that by using the above distance function, we are optimizing robot trajectories to align with the given human demo (i.e. when the human is $p\%$ into the demo, the robot also aims to be roughly $p\%$ into the execution). 

\subsection{Few-shot Motion Optimization with BO}
\label{method:bo}

\edit{With the learned keypoint representation, we can use the keypoint-based distance between a robot execution and a human demonstration as the objective for our optimization problem formulated in Section \ref{sec:prelim}.} 
The remaining problem is how to efficiently find trajectories to be executed on the robot that best imitate the human demo. Training robot skills on large amounts of simulated data and then using sim2real techniques to transfer the skill to a real system is a common robot learning paradigm. However, deformable and granular objects are hard to simulate realistically and therefore pose a challenge for sim2real transfer. Thus, we propose a method to directly optimize the motion policy on the real robot.

\begin{figure}
\vspace{6pt}
    \centering
    \includegraphics[width=0.47\textwidth]{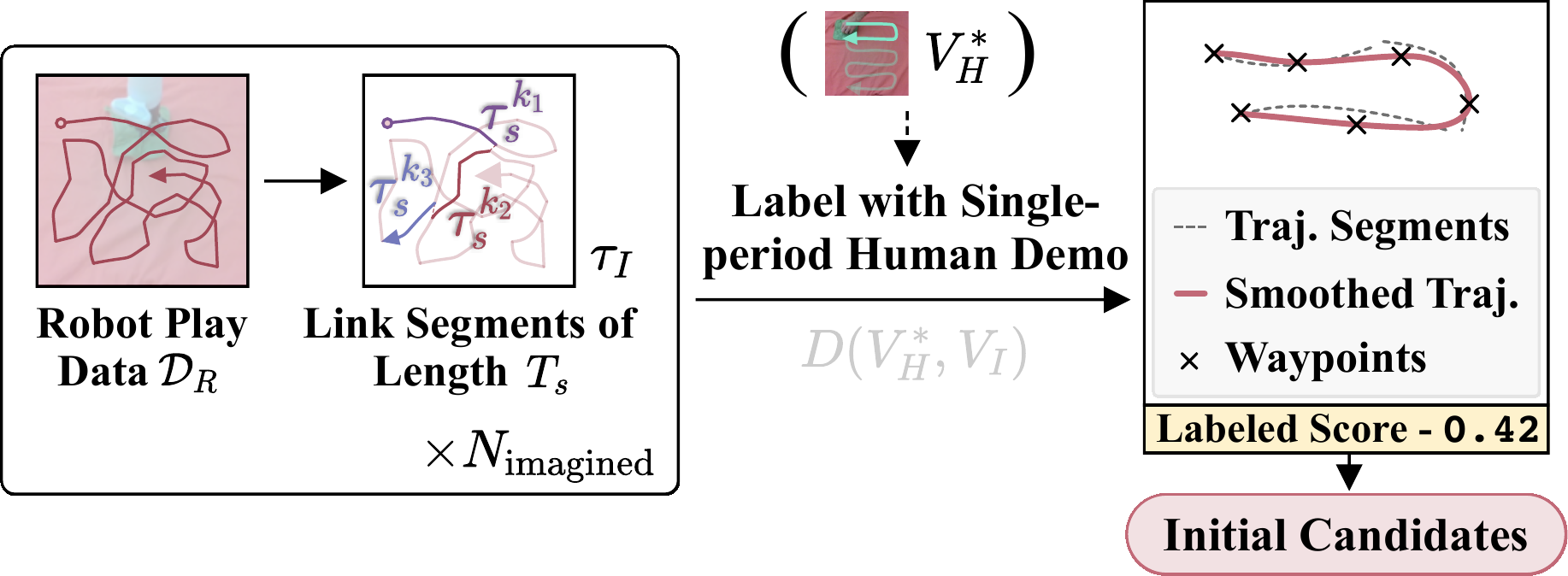}
    \caption{Imagined trajectories as initial candidates for BO.}
    \label{fig:imagined_trajectories}
\end{figure}

\subsubsection{Periodicity Estimation with RepNet}

To imitate periodic manipulation skills shown in the human demo, we need to first determine the periodicity of this demo.
We use RepNet \cite{dwibedi2020counting} -- an approach that can estimate when and how often a periodic task is repeated in a video to estimate periodicity. We observe that the RepNet model trained on the Countix dataset can reliably predict the periodicity of the human demos that we consider. So, we use the trained model (without any finetuning) to predict the number of periods $n^{rep}_H = \text{RepNet}(V_H)$ of the human demo $V_H$.

\begin{figure*}[h]
\vspace{6pt}
\centering
    \begin{subfigure}{.28\textwidth}
      \centering
      \includegraphics[width=\linewidth]{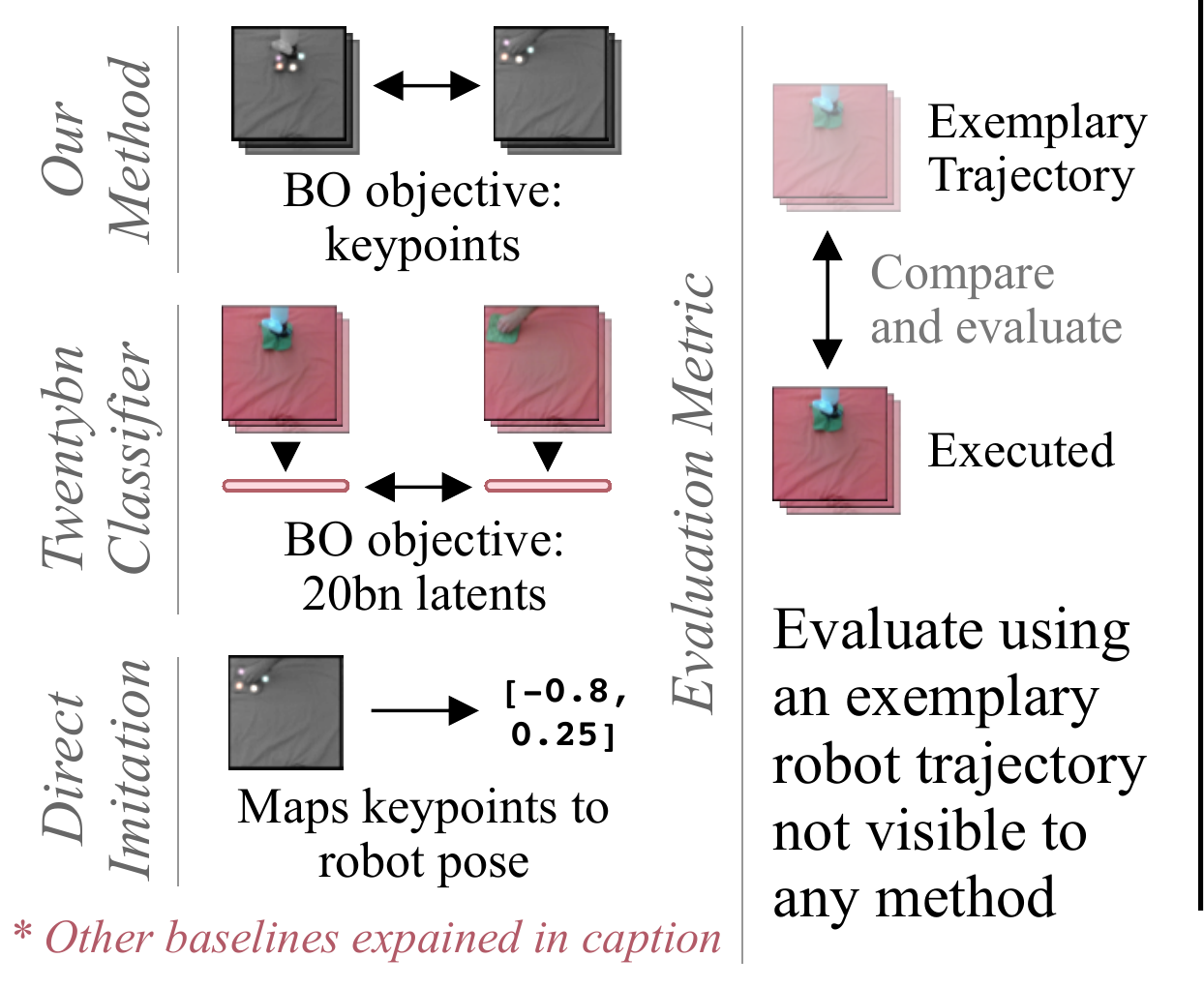}
    \end{subfigure}
    \hspace{1pt}
    \begin{subfigure}{.21\textwidth}
      \centering
      \includegraphics[width=\linewidth]{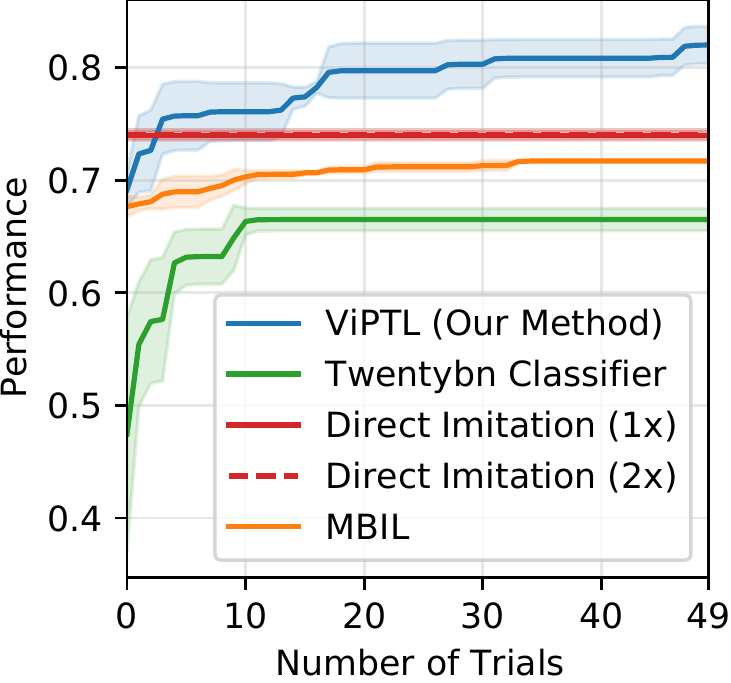}  
      \caption{\texttt{Table Wiping}}
    \end{subfigure}
    \begin{subfigure}{.21\textwidth}
      \centering
      \includegraphics[width=\linewidth]{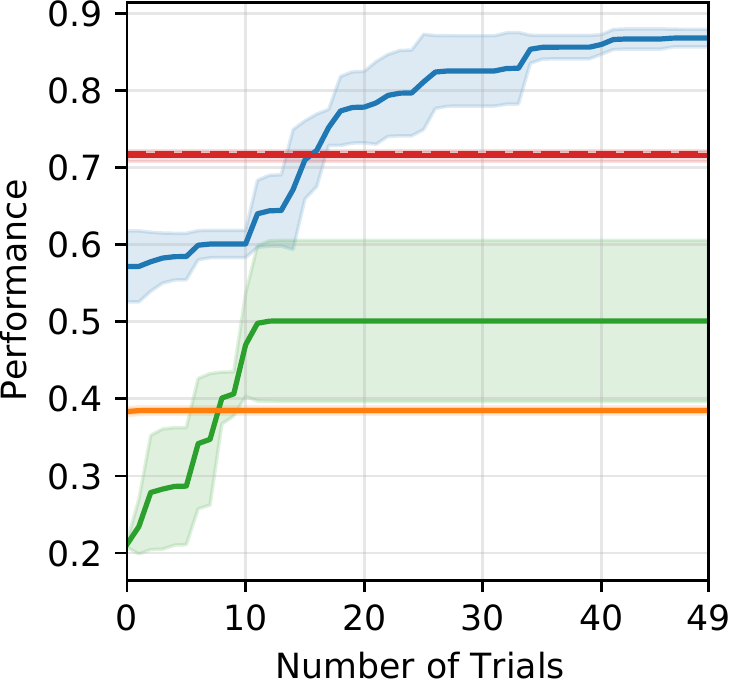}  
      \caption{\texttt{Rope Winding}}
    \end{subfigure}
    \begin{subfigure}{.21\textwidth}
      \centering
      \includegraphics[width=\linewidth]{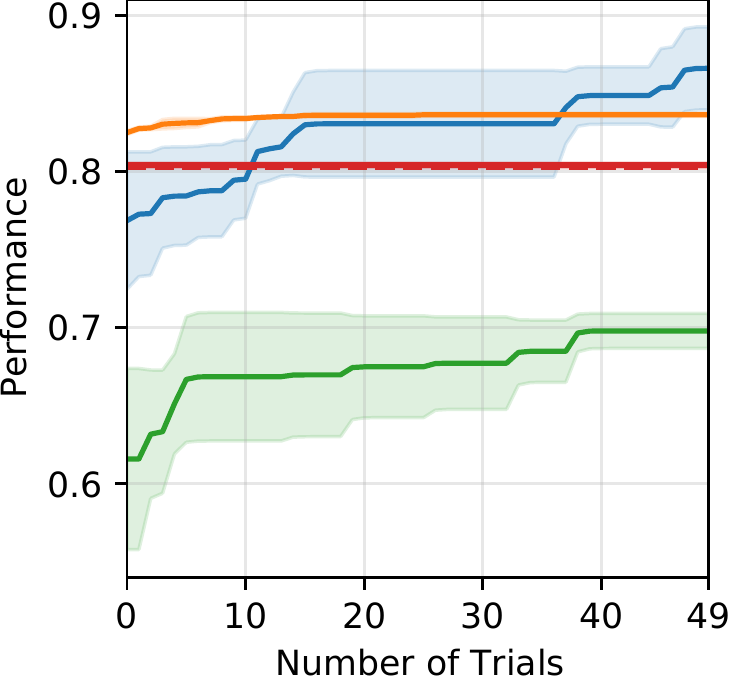}  
      \caption{\texttt{Food Stirring}}
    \end{subfigure}
\caption{
    The left part shows the objectives used in the methods we compare, then explains our evaluation metric -- `performance' in plots on the right. Plots (a)-(c) show the performance of the competing methods in the 3 tasks we consider, in simulation.
    For the \textit{Direct Imitation} baseline, executions are fixed, no finetuning between trials. 
    We include two versions of this baseline: (1x) -- trained from the robot play data; (2x) -- trained from twice the amount of that data, so that the total size of training data exceeds the size of robot play data + the 50 trials of interactions. The performance saturates, showing no benefit from additional training data (red lines match).
    For the other 3 methods, we execute 50 trials for each run. For every method, we do 3 runs using 3 random seeds. The solid lines denote the mean of performance across 3 seeds; the shaded areas denote the standard deviation of the performance values.
    The MBIL line denotes model-based imitation learning baseline, which learns dynamics using keypoints as states and uses MPC for planning.
}
\label{fig:sim_plots}
\end{figure*}

\subsubsection{Motion Optimization with BO}

We propose to optimize single-period waypoints as BO parameters, then use rhythmic DMPs to fit a smooth trajectory and unroll it for multiple periods, as illustrated in Figure~\ref{fig:bo}.
Concretely, BO optimizes single-period waypoints: $\wv = [\vv_1, \vv_2, \ldots, \vv_L]$.
To execute a BO sample, we apply a cubic smoothing to the sampled waypoints, then fit a rhythmic DMP to this smooth trajectory and execute it for $n_H^{rep}$ periods, with the goal $g$ of the DMP shifting $\vv_L - \vv_1$ between two consecutive periods.

During conventional BO, the candidates for the first few trials are sampled at random. In the subsequent trials, an acquisition function samples $N$ candidates at random from the search space, then evaluates their posterior mean and variance to select the most promising next candidate. However, in high-dimensional spaces (above 10D) it is unlikely to sample a well-performing candidate randomly. Even with waypoints as the search space for BO, the space is very large. A leading BO method that recently reported success in high-dimensions~\cite{eriksson2019scalable} was not able to reliably succeed on our tasks within 50 trials. Hence, \edit{we need to} further improve the data efficiency of BO. 
Our insight is that robot play data contains meaningful interactions that can help BO to focus on the promising regions of the search space.
To effectively use this data, we first generate a set of $N_s$ play data segments $\mathcal{S} = \{\tau_s^1, \tau_s^2, \ldots, \tau_s^{N_s}\}$, where each element $\tau_s^i = \edit{\vc{x}}_{RP}^{t_i:t_i + T_s}$ is a randomly sampled fixed length trajectory in the robot play data of length $T_s$. Then, we generate `imagined trajectories' by rejection sampling segment sequences $\tau_I = (\tau_s^{k_1}, \ldots, \tau_s^{k_m})$ such that the end position $\tau_s^{k_i, T_s}$ of each segment $\tau_s^{k_i}$ is less than $d_\text{seg}$ away from the start position $\tau_s^{k_{i+1}, 1}$ of the next segment $\tau_s^{k_{i+1}}$ in L2 distance.
Then, we can find the corresponding image frames of $\tau_I$ to construct an `imagined' video $V_I$,
and this video can be evaluated using the objective score function $D(V_H^*, V_I)$, where $V_H^*$ is a single-period demo trimmed from the original human demo according to RepNet period split. We then select the top $N_{\text{imagined}}$ trajectories with the highest estimated scores and construct a set of \textit{initial candidates}: $\{\wv_i\}_{i=1...N_{\text{imagined}}}$. Each candidate $\wv_i$ is represented by a set of waypoints sub-sampled from the imagined trajectory.
We illustrate this process in Figure~\ref{fig:imagined_trajectories}.

To warm-start BO, we sample the candidates for the first few trials from $\{\wv_i\}_{i=1...N_{\text{imagined}}}$. By construction, these represent the waypoints of the trajectories that have a high alignment with the human demo for the 1st period. In the subsequent trials, we augment the pool of the candidates considered by the acquisition function by sampling in the regions close to these initial candidates. With that, the acquisition function can help us focus the search on the regions close to the initial candidates, but is not restricted to these regions. Hence, we avoid placing hard restrictions on the search space of BO based on prior information. As a result, our BO extension retains theoretical guarantees of BO, such as consistency and regret bounds.
\section{EXPERIMENTS}

In our experiments, we aim to answer the following questions: (1) does our framework successfully learn to perform periodic tasks from a single human demonstration; (2) does our proposed framework perform better than methods that do not exploit periodicity in the target task; (3) is our proposed keypoint-based representation more suitable for learning from human demonstrations than other representations (e.g. latent vectors generated by 20BN classifiers)?

\begin{figure*}[h]
\vspace{6pt}
\centering
    \begin{subfigure}{.21\textwidth}
      \centering
      \includegraphics[width=\linewidth]{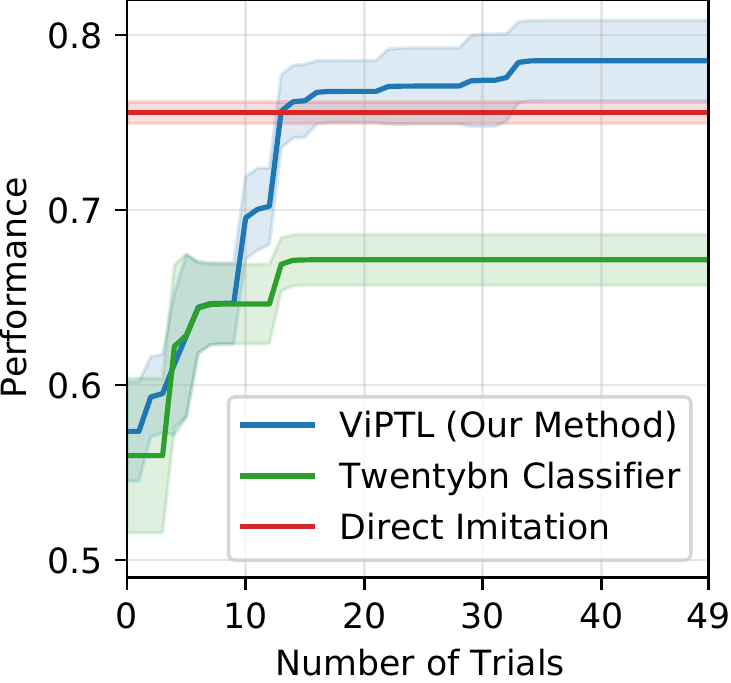}  
      \caption{\texttt{Table Wiping}}
    \end{subfigure}
    \begin{subfigure}{.21\textwidth}
     \centering
     \includegraphics[width=\linewidth]{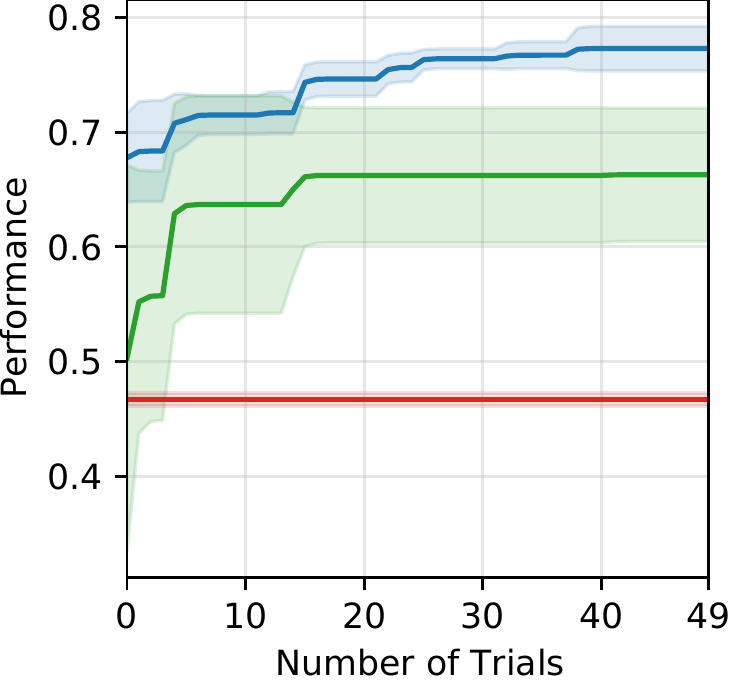}  
     \caption{\texttt{Rope Winding}}
    \end{subfigure}
    \begin{subfigure}{.21\textwidth}
     \centering
     \includegraphics[width=\linewidth]{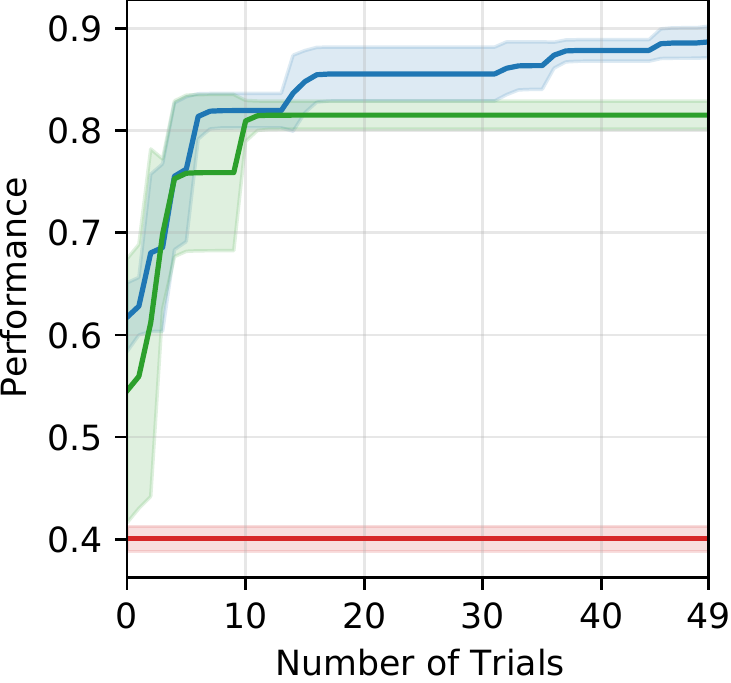}  
     \caption{\texttt{Food Stirring}}
    \end{subfigure}
    \hspace{2pt}
    \begin{subfigure}{.23\textwidth}
      \centering
      \includegraphics[width=\linewidth]{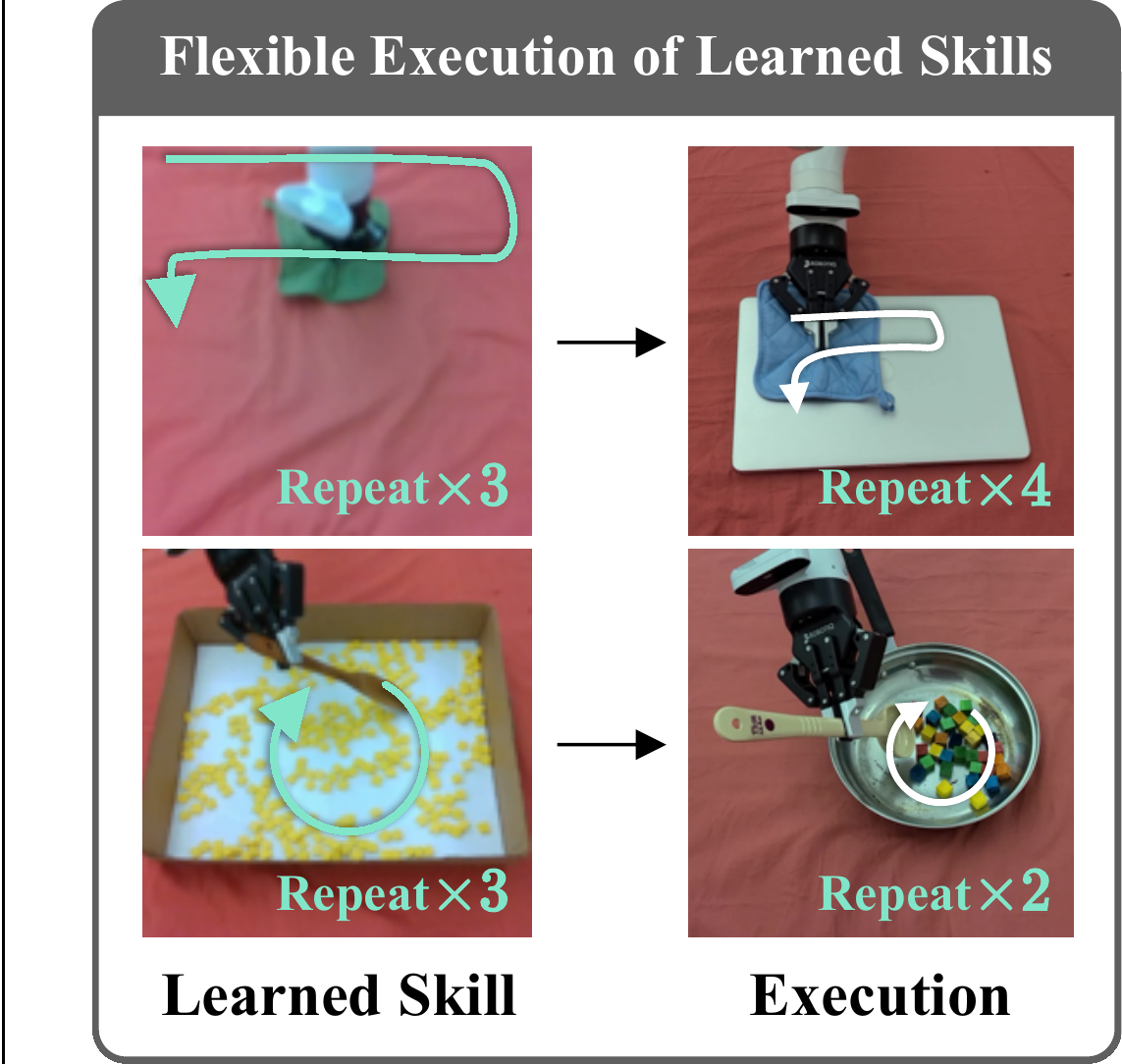}
    \end{subfigure}
\caption{Results for the real robot experiments. Plots (a)-(c) show performance comparisons (for details of the methods see Section \ref{sec:baselines}, for evaluation metric details see Figure \ref{fig:sim_plots}). The right part shows  skills learned with our method executing at different scales and numbers of repetitions without retraining.}
\label{fig:real_plots}
\end{figure*}

We consider 3 challenging manipulation tasks: (1) \texttt{Table Wiping}, where the objective is to wipe a rectangular table surface with a cloth using back-and-forth motions, shifting to cover all the visible the area of the table; (2) \texttt{Rope Winding}, where the objective is to wrap a rope around a fixed spool by repeating circular winding motion several times; (3) \texttt{Food Stirring}, where the objective is to stir granular objects in a tray with a spatula/spoon.

\paragraph*{Metric} The baselines and ablations are each optimizing a different objective function. Therefore, we define a performance metric that is comparable across the different approaches. 
We prepare an exemplary robot trajectory $\tau_E = ( \edit{\vc{x}}_E^1, \ldots, \edit{\vc{x}}_E^{T_E})$ that, when executed, produces the same effect on the objects being manipulated as the human demo. During execution, the robot will produce a trajectory $\tau_R = (\edit{\vc{x}}_R^{1}, \ldots, \edit{\vc{x}}_R^{T})$, where $\edit{\vc{x}}^t$ denotes the robot end-effector position at timestep $t$. \edit{To evaluate the performance of the robot trajectory $\tau_R$, we compute the similarity of $\tau_R$ to $\tau_E$ through $\kappa(-\lVert \tau_E - \tau_{R'} \rVert_1)$,} where $\tau_{R'}$ is a sub-sampled trajectory of $\tau_R$ with length $T_E$ and $\kappa$ is a linear transformation that ensures a score between 0 and 1.
Note that $\tau_E$ is only used for evaluation purposes and not visible to any of the methods. Figures~\ref{fig:sim_plots} and~\ref{fig:real_plots} plot this metric for our simulation and hardware experiments.
The methods that use BO optimize different objective functions e.g. the keypoint-based objective for \textit{\name~(our method)} or the cosine distance between latent features for the \textit{Twentybn Classifier}.

\subsection{Simulation Experiments}

\subsubsection{Baselines and ablations} \label{sec:baselines}
As appropriate baselines and ablations, we need methods that can imitate a single visual human demonstration on the robot. Standard image-based model-free and model-based RL methods \cite{hafner2020mastering, kalashnikov2018qt, ebert2018visual, chebotar2021actionable} cannot operate in this setup because images from human demos and robot execution are visually different. Thus, we use two baselines that use our keypoint-based visual representation as state and an ablation that does not use this visual representation to test whether our visual representation contributes to the final performance of our method.

First we have \textit{Direct Imitation}, which learns a function that maps keypoints at the current timestep to desired robot end-effector positions. This function is trained on robot play data, which contains both keypoints and robot trajectories. To imitate a human demonstration, we use keypoints from the demo video frames as input and output desired robot end-effector positions. The resulting robot trajectory is then executed by fitting a DMP to the predicted robot positions. This baseline studies the use of BO versus a trained neural network for optimizing DMP controllers.

Second we have the \textit{Twentybn Classifier}, an ablation in which the BO objective is based on the video activity classifier~\cite{wang2021tdn} trained on the 20bn dataset~\cite{goyal2017something}. Both the human demonstration and robot execution video are input to the classifier to obtain two feature vectors from the last hidden layer. The objective function is the cosine distance between these two features. We use this baseline to test if the keypoint-based visual representation is a suitable visual representation for the learning from human demo setup compared to alternatives.

Third we have \textit{MBIL}, a model-based imitation learning baseline that relies on a learned dynamics model of the keypoints. The model is trained on robot play data and updated every episode as new interaction data is collected. This model is then used to plan actions to imitate the human demo. At every timestep, we run single-step model-predictive control (MPC) by sampling 5,000 random actions and executing the action that leads to the smallest keypoint distance to the corresponding human demo frame. This baseline tests if our method outperforms model-based RL methods like \cite{manuelli2020keypoints} that do not model periodicity of the task.

\subsubsection{Experimental Setup}

For all our experiments, we use an image size of $512 \times 512$. In the distance function, the number of sub-sampled frames $N_s$ is set to $10\cdot n_H^{rep}$.
To create masks for the human hand, we use the MediaPipe library \cite{lugaresi2019mediapipe} to detect the hand skeleton from an image frame and use the color at the joints of the skeleton to construct a color range mask for the hand. We mask out the robot based on the depth readings (since the robot pose is known).
In BO, we optimize L=7 trajectory waypoints for wiping and winding, L=5 for stirring. 
We use UCB~\cite{srinivas2009gaussian} with $\beta = 0.1$ in the acquisition function of BO. We use automatic hyperparameter optimization to find the appropriate length scales of the RBF kernel. 
When constructing imagined trajectories, we use $10$ play data segments of length $T_s = 10$ each and use a distance threshold of $d_\text{seg} = \frac{1}{6}\lambda_\text{disp}$. We generate 5,000 imagined trajectories and select the top $N_{\text{imagined}} = 100$ trajectories as initial candidates for BO, and use 10 of these in the first 10 BO trials.

\subsubsection{Quantitative Results}

To evaluate the performance of our framework in comparison to competing methods, we run all methods and baselines for 50 trials in all tasks using 3 different random seeds. The performance of all the methods during BO trials is shown in Figure \ref{fig:sim_plots}. The \textit{Direct Imitation} and \textit{MBIL} baselines cannot imitate the human demo well in \texttt{Table Wiping} and \texttt{Rope Winding}, since modeling or capturing dynamics of the deformable objects in these tasks is difficult. 
The suboptimal performance of these two baselines shows that our method outperforms methods that rely on single-step predictions or learning accurate dynamics models and do not exploit periodicity in the target task. 
The \textit{Twentybn Classifier} baseline achieves limited performance in all three tasks as it lacks precision in the classifier-based distance metric used to compare the given human demo with robot executions. This shows that the keypoint-based cost function is a crucial component of our method that is more suitable for the learning from human demos setup. In contrast, our method (\textit{\name}) \edit{achieves good performance in all three tasks, outperforming various baselines and ablations.} 
We include qualitative results in the supplementary video.

\subsection{Real Robot Experiments}

\subsubsection{Hardware Setup}

Our hardware setup (Figure \ref{fig:real_plots}) includes a Kinova Gen3 robot arm with a Robotiq 2F-85 gripper and an Intel RealSense D435 camera. 
The table workspace measures $50 \times 43~\text{cm}$, and the camera is mounted at the height of 68 cm at one side of the workspace.
The camera provides the RGB image data during experiments and is positioned to view the table surface. We use velocity control in the Cartesian (end-effector) space to execute the desired trajectories on the robot.

\subsubsection{Quantitative Results}

We select the \textit{Twentybn Classifier} ablation and the baseline with the most consistent performance across tasks in simulation experiments (\textit{Direct Imitation}) to compare with our method (see Figure~\ref{fig:real_plots}). Our method is able to quickly find high-scoring points due to an effective warm-start and fine-tune the generated motion, leading to consistently improving performance throughout the 50 trials. The baselines are unable to catch up with the performance of our method for similar reasons as mentioned in simulated experiments. \edit{In particular, the \textit{Direct Imitation} baseline performs worse here than in simulation since dynamics in the real world is more stochastic.}
\vspace{2pt}
\section{CONCLUSION}

We introduced Visual Periodic Task Learner (\name), a method for representing periodic manipulation policies and efficiently learning them from a single human demonstration. We show that \name~succeeds on three robot manipulation tasks that involve deformable and granular objects. This work demonstrates the benefit of leveraging the periodic structure of many tasks commonly seen in everyday life. In future work, we intend to extend our method to handle initial stages of tasks, such as grasping and other transient motions. 
\edit{We also plan to extend our perception module so that it generalizes to unseen camera views and contexts.}

\bibliographystyle{references/IEEEtran}
\balance
\bibliography{references/references}

\end{document}